\documentclass{amsart}
\usepackage{diagrams}

 \newtheorem{thm}{Theorem}[subsection]

 \newtheorem{prop}[thm]{Proposition}
 \theoremstyle{definition}
 \newtheorem{defn}[thm]{Definition}
 \theoremstyle{remark}
 \newtheorem{rem}[thm]{Remark}

\begin{document}
\title[Memory as a Monadic Control Construct]
 {Memory as a Monadic Control Construct \\in Problem-Solving}

\author{ Jean-Marie Chauvet }

\address{Dassault D\'{e}veloppement, Paris, France}

\email{jmc@neurondata.org}



\date{ August 4, 2003.}

\dedicatory{}


\begin{abstract}
Recent advances in programming languages study and design have
established a standard way of grounding computational systems
representation in category theory. These formal results led to a
better understanding of issues of control and side-effects in functional and imperative
languages. This framework can be successfully applied to the investigation of 
the performance of Artificial Intelligence (AI) inference and cognitive systems.  In 
this paper, we delineate a categorical formalisation of memory as a control 
structure driving performance in inference systems.  Abstracting away control 
mechanisms from three widely used representations of memory in cognitive systems 
(scripts, production rules and clusters) we explain how categorical triples 
capture the interaction between learning and problem-solving.
\end{abstract}

\maketitle

\section{Reminding as computation}
What are the basic issues in the design of memory systems? Memory
systems must have the ability to cope with new information. Any new
input that is to be processed by a memory system should cause some
adjustment in that system. Memory systems are the quintessential
\emph{state} semantics. A dynamic memory system is one that is altered
by every experience it processes, and, in addition, it must be capable
of finding what it has in it. Retrieval of memorised information, for
instance through reminding, is a of course a critical feature of the
system. More specifically the role of reminding in the overall
architecture is the critical design issue. To be reminded of something
we must have come across it while we were processing the new
input. But to have done so, the memory system either had to be
\emph{looking} for this reminded event or else it must have \emph{run
into it accidentally}. Following Schank and Abelson's seminal work \cite{scripts},
reminding relies on an amalgamation of both scenarii: a memory system
is not consciously looking for a particular episode in memory, because
it doesn't explicitly know of that episode's existence; it knows
however where episodes like the one it is currently processing are
likely to be stored. Reminding then occurs when the memory system has
found the most appropriate structure in memory that will help in
processing a new input. When no one episode is that closely related to
an input, we can still process it, but no reminding occurs. This
implies that an \emph{expert} memory system is constantly receiving
new inputs and evalutating them in terms of previously processed
inputs. It understands in terms of what it already understood.

Schank's original thesis is that since reminding may not necessarily
bring back the most relevant prior experience in processing a new
input, reminding, which occurs naturally during processing by a
dynamic memory system, and processing itself ought to amount to
different views of the \emph{same} mechanism. Let's denote this as the
\emph{reminding as computation} thesis. This thesis can be traced in
several approaches to the implementation of memory sytems and machine learning.

\subsection{Schank and Abelson's scripts}
First and foremost, further research work by Schank and Abelson and
then Schank studied the use of \emph{scripts} as fundamental dynamic
memory structures \cite{scripts,DM}. A script is a collection of specific memories
organised around some common points. A script is built up over time by
repeated encounters with a situation. When an event occurs for the
first time it is categorised uniquely. Repeated encounters with
similar events cause an initial class to contain more and more
episodes. Common elements to those episodes are treated as a unit, a
script. But, subsequent episodes that differ from the script partially
are attached to the part of the script they relate to. The differing
parts of the episode are stored in terms of their difference from the
script. In this way, such episodes can be found when similar
differences are encountered during processing.

In this view mechanisms for aggregating, indexing and processing
information are not separated and occur spontaneously when the system
is confronted with a new situation. A script is a generalisation and
reminding can be seen to be occurring when an expectation is made, but
fails to materialise, or, when something happened that was not
expected and then happens. Indexed under either of these two kinds of
failures are prior failures of the same kind. This view therefore puts
forward \emph{failure-driven} or \emph{expectation-driven} memory
systems. It is based on a set of structures in memory that generate
predictions for use in processing---understanding. The latter are
central to a scheme of storing and finding memories based on
prediction failures.

\subsection{Clustering: Dynamic memory in the NXP architecture}
The goal investigation and evocation mechanisms central to the NXP
architecture are complemented by an interactive dynamic
memory. Following Schank's original thesis and intuition, reminding
and processing inputs in the NXP architecture are but two aspects of
the same cognitive architecture. Inspired from its original
investigation of a specific application domain, clinical medicine, the
dynamic memory system of the NXP architecture helps exploring the impact of
growing experience in the transformation of novice to expert. Numerous
approaches to the question of relating performance to experience have
been investigated, ranging from statistical methods to neural networks
and self-orgnising production systems. We will visit some of them
later at we focus first on the somewhat more modest objective of the
NXP dynamic memory.

Keeping with the thesis of reminding as computation, the NXP dynamic
memory is adjusted with each new situation encountered and
processed. Episodes (in Schank's definition) are constituted of
successive problem-solving sessions. Each episode is then a complete,
possibly interactive, evaluation of goals relevant to a particular
situation in the environment---the evaluation itself involving
arbitrary sequences of goal investigations and evocations in the sense
defined and formalised in \cite{Rappaport_1984_77}. The episodes are constantly added
to the NXP dynamic memory by aggregation resulting in a hierarchy of
clusters of \emph{similar} problem-solving behaviours. Clusters of
behaviours are comparable to Schank's scripts or MOPs (\emph{memories
organised packet}). In the NXP architecture clusters are used to
produce \emph{expectations} influencing further problem-solving
activity. The first investigation, conducted in the application domain
of clinical medicine \cite{Rappaport_1984_77}, presented qualitative results relating
performance of the problem-solving process, viewed as search in the
goal space, to an initial expectation derived from the clusters of the
dynamic memory. In that study, the expectation, in the form of
\emph{first-look signs}, was used to evoke an initial set of goals for
investigation. Even with this simplified definition of medical
expertise (predicated on the sole ability to \emph{pre-structure} the
problem uniquely as a result of the compilation of personal experience,
ignoring other abilities or learning by being taught for instance), the NXP
architecture demonstrated acquisition of expert behaviour. Accuracy of
the expectation, measured in terms of the relevance of the evoked
goal, indeed increased as new (expected and unexpected) cases were
presented to the system. In particular, the first occurrence of an
unexpected situation provokes adjustment of the clusters hierarchy
allowing a rapid, efficient recognition of further occurrences later
in time.

\subsection{Chunking : learning from impasses in problem-solving}
Another fundamental proposition for dynamic memory architecture was
put forward by Newell \cite{Soar}. Embedded into the Soar architecture,
three major design assertions capture much of the processes of
reminding and learning.

The first assumption is the functional unity of long-term memory
(LTM): all long-term memory consists of recognize-retrieve units with
a fine grain of modularity. (In Soar these are production rules.) It
follows that episodic, semantic, and procedural knowledge must be all
encoded in this same dynamic memory---a tenet of the \emph{unified
theories of cognition} view. How those abstractly posited structures
are actually implemented in the neural circuitry is still an open
question, but the body of experimental results informally tend to
support this hypothesis. The second assumption is the
learning-by-chunking hypothesis: all long-term learning occurs by
chunking. The last assertion is the functionality of short-term or
working memory: short-term memory arises because of functional
requirements. It is basically the locus of interaction between the
various consituents of the architecture.

Chunking can be viewed as a form of permanent goal-based caching. In
contrast to more traditional production or inference systems, the Soar
architecture creates subgoals dynamically when it runs into
impasses. If Soar knows what to do next, there is no need of a
subgoal. The subgoal to resolve the impasse is created when the
decision procedure---the current goal evaluation---can not choose a
unique decision. Once the impasse is resolved, a new LTM element is
created from the working memory elements that existed prior to the
impasse and were found to give rise to the result. Chunking is a
continuous process and applies to all subgoals whatever their level of
abstractions: it is thus a so called \emph{weak method}\cite{Soar}. Indeed Soar
experiments displayed learning by chunking of search control, operator
implementation, creation of new problem spaces, etc.

\section{Formal aspects of dynamic memory semantics}
In Schank and Abelson's script-based dynamic memory, side-effects of script execution may result in amendments or creation of new scripts.  In clustering-based cognitive systems, such as the NXP dynamic memory, the classification of new input may trigger a reorganisation of the cluster hierarchy itself. Similarly in Soar, chunking, triggered by the problem solving process may result in amended or new production rules in the knowledge base. The common theme of new or amended control constructs as side-effects resulting from control execution itself has been widely studied in the context of imperative programming languages.

The idea of transforming programs to
continuation-passing style (CPS) appeared in the mid-sixties
\cite{sabry92reasoning}. The transformation was formally codified
by Fischer and Reynolds in 1972, yielding a standard CPS
representation of call-by-value lambda calculus. In the context of
denotational semantics \cite{Stoy}, described as the theory of
meaning for programs, denotations are usually built with the help
of functions in some mathematical structures. In CPS these
functions are explicitly passed an additional argument $\kappa$
representing "the rest of the computation", a first step towards
full reification of the notion of computation itself. Although
most of the work on continuations involves a functional language,
usually a fragment of typed or untyped lambda calculus, CPS is
also useful in understanding imperative languages. Continuations
were found to be a major tool for the design of interpreters 
and compilers for many languages, most prominently Scheme, ML and Haskell.

As such, continuations appear as the raw material of control.
Operations on continuations control of the
unfolding of a computation---that this translation happens in a standard way
is a major result of the theoretical work on continuations.
Similarly, in inference systems, of which production systems are a
well-studied example \cite{PDIS}, designing a proper behaviour
relies on the delicate interweaving of \emph{goal-driven} and
\emph{data-driven} control.

\subsection{Continuation-passing style transformations in imperative and functional programming languages}
Practically all programming languages have some form of control
structure or jumping; the more advanced forms of control
structures tend to resemble function calls, so much so they are
rarely describes as jumps. The library function {\tt exit} in C,
for example, may be called with an argument like a function. Its
whole purpose, however, is utterly non functional: it jumps out of
arbitrarily many surrounding blocks and pending function calls.
Such a non-returning function or jump with arguments is an example
of why continuations are needed.

In  continuation-passing style, a function call is transformed
into a jump with arguments to the callee such that one of these is
a continuation that enables the callee to jump back to the caller.
This idea has been formalised into a standard CPS transformation
for lambda calculus. 
\begin{defn} \label{d:Lambda-calculus}
The pure untyped lambda calculus $\Lambda$ is defined by \cite{
Barendregt-1997}: A set of terms, $M$, inductively generated over an
infinite set of variables $Vars$,
\begin{itemize}
\item Terms $M ::= V |\: M\: M $; \item Values $V ::= x |\:\lambda
x.M $,  $x$ in $Vars$.
\end{itemize}
\end{defn}

The standards CPS translation for $\Lambda$ in denotational
semantics is given by the map $[[\_]]_{F} : \Lambda \rightarrow
\Lambda$ originally described by Fisher:
\begin{defn} \label{d:FisherCPS}
Let $k$, $m$, $n$ in $Vars$ be variables that do not occur in the
argument to $[[\_]]_{F}$.
\begin{itemize}
\item $[[V]]_{F} = \lambda k.k \psi (V)$ \item $[[M N]]_{F}
= \lambda k.[[M]]_{F}(\lambda m.[[N]]_{F} \lambda n.(m k)n)$ \item
$\psi (x) = x$ \item $\psi (\lambda x.M) = \lambda k.\lambda
x.[[M]]_{F}k$
\end{itemize}
\end{defn}

The continuation-passing style puts foward composition as a basic building block in the construction of programs' semantics.  Composition is in essence a categorical notion.  Moggi \cite{moggi91notions} originally introduced the use of category theory as a mathematical tool to study the semantics of programming languages, sparkling a whole new approach to formalising computational processes \cite{wadler93monads, sabry92reasoning, filinski96controlling}.  Because composition is the recurring basic process in dynamic memory scripting, clustering, and chunking the monad, or triples, a categorical construction is a good candidate for the formal representation of dynamic memory in its interactions with problem-solving.

\subsection{Script-building as composition}
A CPS semantics for Schank and Abelson's script style dynamic memory requires we establish the distinction between  scripts in the agent's memory, tasks in the agent's environment and (possibly partial) \emph{executions} of tasks.  With $S$ the set of scripts and $T$ the set of tasks, both further unspecified in the scope of this first semantics denotation, a task execution is captured by a functor $k: (t, s)  \rightarrow k\: t\: s$ from $T S \rightarrow O$, where $O$ is some suitable domain of output objects.  The semantics function for the dynamic memory operation can then be defined as follows.

\begin{defn} \label{d:CompositionScripts}
For a given script $s$ in $S$, we introduce the execution semantics function $[[ \_ ]]\: s\: k : T \rightarrow T S (T S  \rightarrow O)$.  The expression $[[t]]\: s\: k$ simply denotes the execution of task $t$ with reference to script $s$, $k$ being the \emph{continuation}, i.e. the pending agent's task execution.
\end{defn}
With definition \ref{d:CompositionScripts}, the interweaving of task execution and knowledge acquisition ---the alteration or creation of scripts as a side effect of task execution itself---is reflected by considering the sequence operator "$;$" $ : T T \rightarrow T$ which defines $t_{1} ; t_{2}$ as the task constituted of the performance of task $t_{1}$ followed by the performance of task $t_{2}$ and its associated semantics:
\begin{defn}\label{d:CompositionScriptSequence}
For $t_{1}$ and $t_{2}$ both in $T$, $s$ in $S$ and $k$ a continuation, $$[[t_{1};t_{2}]]\: s\: k = [[t_{1}]]\: s\: (\lambda \sigma . [[t_{2}]]\: \sigma \:k)$$ which expresses that the execution of $t_{1}$ with script $s$ is passed to the continuation constituted by the execution of task $t_{2}$ with the script (altered or new) resulting from $t_{1}$ execution, followed by resuming with the original continuation.
\end{defn}
Definitions \ref{d:CompositionScripts} and \ref{d:CompositionScriptSequence} oversimplify several features of the original proposition by Schank and Abelson.  They consider only one script whereas a collection of scripts is in fact available to the agent, hence they ignore amendments altering the organisation of scripts rather than individual scripts.  In addition both definitions leave unspecified the details of the knowledge acquisition process itself; the failure-driven nature of script alteration, essential to Schank and Abelson's model, is out of their scope. Modelling the nature of the knowledge acquisition process requires another layer of semantics denotation concerned with the explicit failure of the current script execution---a distinguished {\tt fail} continuation in \ref{d:CompositionScripts}'s model---and its mapping into the set $S$ of scripts, a functor $KA: T S (T S \rightarrow O) \rightarrow S$ with specific denotation for $KA(t,s,{\tt fail})$.

\subsection{Clustering as composition}
In a clustering-based cognitive system, the goal acquisition process depends on \emph{clusters} of instances of previous problem-solving activities.  Memory is structured as a hierarchical mesh of such clusters, in a way not unlike scripts in Schank and Abelson's model.  The dynamic nature of clustered memory stems from the modifications to these hierarchies entailed by the problem-solving activity.

In the NXP Architecture, for instance, each path to the solution of a newly solved problem is input to a classifier producing a hierarchy of clusters incorportating the new instance together with the previous ones.  The resulting cluster hierarchy is then tapped for a set of initial goals, called \emph{first-look signs}, blending a goal-driven process with the data-driven one of the next problem-solving episode. In contrast to chunking or script enacting, clusters are only used in the initial phase of problem solving, setting up the background context in which the agent is to execute the task.  In chunking and in script enacting, new chunks or new scripts are added on the fly while the agent progresses toward a solution and become immediately available.  The NXP Architecture sports a different mechanism for goal generation during problem solving, the so called \emph{goal evocation} \cite{chauvet2002}.

In line with definition \ref{d:CompositionScripts} we introduce:
\begin{defn}\label{d:CompositionClusters}
Let $H$ be the set of clusters hierarchies (a boolean algebra, the elements of which are powersets of task features), $T$ the set of problem-solving tasks (a task being considered here as a set of task features), and $k$ a continuation in $T H \rightarrow O$ where $O$ is a suitable domain of output objects, ";" being, as before, the task concatenation operator, the operational semantics of clustering is defined by a functor $[[ \_ ]]\:h \:k \rightarrow T H (T H \rightarrow O)$ for which $$[[t_{1};t_{2}]]\: h\: k = [[t_{1}]]\: h\: (\lambda \delta . [[t_{2}]]\: \delta \:k).$$
\end{defn}
As in definition \ref{d:CompositionScriptSequence}, the latter equality expresses that solving a sequence of problems involves solving the first one and passing control to the next problem-solving activity in the context of the cluster hierarchy modified by the first problem-solving activity.  As in the previous subsection, definition \ref{d:CompositionClusters} somewhat simplifies the agent's cognitive processes.  The nature of the changes to cluster hierarchies is left unspecified as are the criteria used for classifying the solved problem.  This definition neither considers the goal evocation mechanism of the NXP Architecture nor on-the-fly clustering and goal extraction during problem solving which could be added to the solving engine.  In fact, a formal model of the latter would be similar to the next subsection model of chunking.
   
\subsection{Chunking as composition}
In contrast to the initial goal setup in the NXP Architecture, Soar's chunking alters long term memory by adding production rules to the agent's knowledge base.  These new production rules become instantly available and can be brought to bear later on the current instance of problem-solving. This \emph{learn-as-you-walk} style of search, like Schank and Abelson's dynamic memory, is driven by the ability to recognise and resolve \emph{impasses}, possibly generating new problem-solving activities on-the-fly.

These similarities lead us to propose the following composition semantics for chunking:
\begin{defn}\label{d:CompositionChunks}
Considering $T$ the set of tasks, $R$ the powerset of production rules, and continuations as functions $T R \rightarrow O$ into a suitable domain of output objects, the semantics functor for chunking-based systems is $[[\_]]\:r \:k : T \rightarrow T R (T R \rightarrow O)$ with, as before, the following equality: $$[[t_{1};t_{2}]]\: r\: k = [[t_{1}]]\: h\: (\lambda \rho . [[t_{2}]]\: \rho \:k).$$
\end{defn}
As in the previous subsections, this definition leaves a number of important features unspecified: the process through which impasses are actually resolved, the integration of newly generated rules in the long term production memory, conflict resolution when impasses similar to previously encountered situations occur during further search, and so forth. Our operational semantics only attempts to capture the knowledge acquisition process as a side-effect of the problem-solving process---borrowing from the literature and results on formalisation of side-effects in programming languages.

Because of this somewhat simplified view of the close weaving of knowledge application and elicitation in AI inference systems, ignoring specificities which are nonetheless important and often critical in the cognitive models embedded in each system under consideration, we are led to a \emph{common} abstract view of the integration of knowledge acquisition and performance systems.  This intuition of the existence of a common abstraction, detailed and formalised in the next section, underlies the readily apparent similarities in definitions \ref{d:CompositionScriptSequence}, \ref{d:CompositionClusters}, and \ref{d:CompositionChunks}.

\subsection{Simplified continuation semantics for knowledge acquisition}
Without loss of generality we will formalise an inference system---a cognitive system performing problem-solving tasks---as a \emph{deterministic transition system}.
\begin{defn}\label{d:DTS}
A deterministic transition system $T$ is a 7-uple $\left\langle C, I, F, \Omega, D, \alpha, S \right\rangle$ where the set $C$ of configurations is the disjoint union of $I$, $F$ and $\{\Omega\}$, respectively a set of initial configurations, a set of final configurations and $\Omega$ the undefined configuration; $D$ is a suitable domain of values, $\alpha: F \rightarrow D$ a valuation assigning a value to each final configuration and $S$ a deteministic \emph{step}, or transition relation i.e. a partial function $S: C \rightarrow C$ with $dom(S) \subseteq I$.
\end{defn}
A deterministic transition system (DTS) can be seen as an abstract form of a program defined as the sequential execution of problem-solving tasks in an environment.  In this formal framework, the transition relation step represents the execution of one problem-solving task.  The infinite tree obtained by unfolding (or unraveling) a transition system can be seen as its global behaviour and the history of task executions is captured as a path in the unfolding of the associated DTS.  A natural denotational semantics for a DTS can be defined by using continuations.

In order to do so, we first introduce a suitable set of states $\Sigma$, a mathematical entity into which the semantics function will map task executions.  In fact \cite{Stoy} to be a suitable domain, this mathematical entity requires additional properties such as the possibility to define monotonicity and continuity of functions ranging over it.  A \emph{complete lattice} is one of the most simple of such satisfactory structures : a set with an order relation such that each pair of elements has a greatest common lower bound and a single element, usually denoted $\bot$, lower than every other.  Let $\Sigma_{\bot}$ denotes a fixed complete lattice; consider a task $t$ that is part of a larger execution history, the denotation $[[t]]$ will be a function that will, in the end, deliver an answer in $\Sigma_{\bot}$.

The task itself is executed within a particular environment $\eta \in Env$ which defines the symbol $t$ and any of the other subtasks required by the execution of $t$.  The execution of a task applies to a state $\sigma \in \Sigma$ and returns a new state in $\Sigma_{\bot}$ reflecting a stepwise transition in the DTS.  The semantics function is then a function $[[.]]: Env \rightarrow \Sigma \rightarrow \Sigma_{\bot}$, and the semantics denotation of a task $t$ is written as the state $[[t]]\eta\sigma$.

This, however, ignores the flow of control in the history of task execution, i.e. the behaviour.  The final answer is not the result of executing $t$ alone, but the result of evaluating the whole history of which $t$ is a subtask.  In order to capture this forward looking effect, we specify that the denotation additionally depends on the the remainder of the history to be executed once the task is performed.  In some knowledge representation frameworks this remainder which need to be factored in the semantics definition is alternatively called the set of \emph{goals}\cite{Soar, 27702, PDIS}, \emph{postponed evaluations}\cite{chauvet2002}, or \emph{expectations}\cite{DM}.  This leads to the refined functionality $[[.]]: Env \rightarrow Cont \rightarrow \Sigma \rightarrow \Sigma_{\bot}$.  Here the remainder of the execution is a function $\xi: \Sigma \rightarrow \Sigma_{\bot}$ since the future performance will in the end yield an answer.  The semantics denotation of the execution of $t$ now becomes $[[t]]\eta\xi\sigma$.

The semantics of the basic sequence operator ';' $[[t_{1};t_{2}]]\eta\xi\sigma$ says that the answer obtained from executing $t_{1};t_{2}$ is equal to the answer resulting from execution of $t_{1}$ in the continuation composed of the execution of $t_{2}$ followed by the original continuation:
$$[[t_{1};t_{2}]]\eta\xi\sigma = [[t_{1}]]\eta\{[[t_{2}]]\eta\xi\}\sigma$$

The next step is to introduce knowledge acquisition resulting from unexpected situations met during a problem solving task.  Whether in the case of Schank and Abelson's scripts, SOAR or NXP architectures, unanticipated issues trigger a reorganisation of knowledge in long-term memory.  In Schank and Abelson's models of dynamic memory, an unexpected difference in an episode provokes a computation of the difference between the differing event and the class of expected events from the script, and a subsequent alteration of this class in long-term memory with an extension storing the computed difference for later retrieval.  In SOAR, unexpected situations are resolved by, firstly, creating a new problem sub-space for local resolution, and, secondly, chunking which adds the newly found local solution as a new set of general production rules in long-term memory.  In the NXP Architecture, the end of the problem-solving task triggers a recomputation of the clusters in long-term memory to properly classify the new instance.  Knowledge acquisition and reminding are nothing more than another form of computation, somewhat orthogonal to the computation involved in problem solving, but of the same nature and tightly knit with the task execution itself.  In the next paragraphs we try to explicit this orthogonality in a formal way.

\subsubsection{Expected and unexpected continuations}
Describing the flow of control is more complicated now.  The notion of \emph{future of computation} is not that obvious anymore.  Task execution can behave in two different ways now: expected performance, which is similar to what we had before, or meeting of an unexpected situation triggering knowledge acquisition. We capture this double-edged future by two continuations instead of one, an \emph{expected} continuation, $\xi \in ExpCont$, similar in essence to the one introduced in previous paragraphs, and an \emph{unexpected} continuation, $\mu \in UnexpCont$, reflecting that the remainder of the task performance happens with a modified long-term memory.

Returning to the initial presentation of a problem-solving engine as a DTS, $UnexpCont$ can be seen as the set of dynamic changes of states and transition definitions occuring on the fly as the engine unfolds the DTS.  Alternatively, this set of changes can be viewed as a morphism of DTSes, with the engine starting to unravel a DTS only to switch to the image of the original DTS through the morphism as it encounters an unexpected situation.

\begin{defn}\label{d:ContinuationSemantics}
The final form of the semantics function is now $[[.]]: Env \rightarrow ExpCont \rightarrow UnexpCont \rightarrow \Sigma \rightarrow \Sigma_{\bot}$, where the denotation $[[t]]\eta\xi\mu\sigma$ of task $t$ will ultimately depend on $\xi$, the rest of the task sequence following $t$, and on $\mu$ which is a denotation of the rest of the task sequence occuring in the context of a modified long-term memory.
\end{defn}

\begin{rem}
The meaning of an unexpected situation, $[[\tt{unexp}]]$, is straightforward.  The answer is simply the one provided by the unexpected continuation: $$[[\tt{unexp}]]\eta\xi\mu\sigma = \mu$$
\end{rem}

Revisiting the factorisation of sequence of subtasks into taks and histories, we say that a task $t$ is \emph{composed} of tasks $t_{1}$ and $t_{2}$, written $t = t_{1} \circ t_{2}$, when the first one is performed followed by the second one in whichever context the long-term memory is left after the first execution.  Abstracting the general idea from the various dynamic memory models of Schank and Abelson, SOAR and NXP we denote $\phi$ the computation yielding the new long-term memory state---difference matching in Schank and Abelson's, chunking in SOAR, clustering in the NXP Architecture.  The nature of $\phi$ is left unspecified at this stage as its precise form is dependent on the dynamic memory model.  It can only be stated that $\phi$ is a \emph{weak method}\cite{Soar, 27702} and that it is computable.

\begin{prop}\label{p:Composition}
Let $d$ be a DTS representing a cognitive system with weak method $\phi$ for knowledge acquisition, and semantics function as in \ref{d:ContinuationSemantics}.  For tasks $t_{1}$ and $t_{2}$,
$$[[t_{1} \circ t_{2}]]\eta\xi\mu\sigma = [[t_{1}]]\eta \{ [[t_{2}]]\eta\xi \} \{ [[\phi]]\eta\{[[t_{2}]]\eta\xi\}\mu\}\sigma$$

\begin{proof}
The expected continuation $\xi'$ of execution of task $t_{1}$ is the execution of task $t_{2}$ followed by the original expected continuation:
$$\xi' = [[t_{2}]]\eta\xi$$
The unexpected continuation $\mu'$ of execution of task $t_{1}$ is constituted of the execution of task $t_{2}$, also followed by the original continuations (both expected and unexpected as performing $t_{2}$ might introduce unexpected situations as well) but in the context of a dynamic memory modified by $\phi$.
$$\mu' = [[\phi]]\eta\xi'\mu$$
Replacing $\xi'$ and $\mu'$ in $[[t_{1} \circ t_{2}]]\eta\xi\mu\sigma = [[t_{1}]]\eta\xi'\mu'\sigma$ yields the proposition result.
\end{proof}
\end{prop}

\begin{rem}
In the previous definition and proposition the knowledge acquisition process appears orthogonal to the basic problem-solving computation.  The expected and unexpected continuations are separate arguments to the semantics function.  Their side-effects, however, are tightly linked: meeting an unanticipated issue, here computing $\tt{unexp}$, triggers $\phi$ before carrying on further tasks; on the other hand, executing $\phi$ basically changes the long-term memory, or equivalently switches on the fly to another DTS computed from the current one, therefore impacting all further task execution.
\end{rem}
\begin{rem}
The denotational semantics proposed in this section is abstract.  It does not specify an \emph{operational} semantics for a possible implementation of a dynamic memory.  It is, however, useful in forging the general principles along which to map the computations of dynamic memory into a particular execution engine such as a stack machine \cite{chauvet2002}, a generic production system, an imperative or logical programming language.
\end{rem}
\begin{rem}
Looking further into implementation issues, it is apparent from proposition \ref{p:Composition} that the execution of $t_{2}$ appears twice in the computation of a composition in which it is involved.  This raises the issue of optimisation when compiling or interpreting compositions.
\end{rem}

\subsubsection{ Continuations, exceptions and states }
The \emph{bi-continuation} approach outlined in the previous section directly mirrors an explicit separation between actual performance and knowledge acquisition.  This is readily apparent in the introduction of $\phi$, the otherwise undescribed acquisition method, and $\tt{unexp}$ a singled out continuation triggering a memory change according to $\phi$.  This approach may overlook aspects of cognitive systems performance related to the intended \emph{semantics} of the knowledge acquisition process.

In SOAR, for instance, while meeting an unexpected situation immediately provokes the creation of a new problem space, possibly with its own set of states and operators, the problem solving method in the new space, however, is not different from the one used in the original space.  The semantics there is similar to calling a subprogram or handling a failure in a modern programming language.  In this respect, the \emph{unexpected continuation} really behaves more like an exception.  Once the unexpected situation is resolved in the subspace, its solution is brought upward in the original space and execution can resume in this space.  In this respect, the exception handler basically returns control to the original (expected) continuation.  This behaviour does not affect though the notion that $[[t]] \cong [[t \circ t]]$ (\emph{contextual equivalence}) with the intent of stating that solving the same problem twice is broadly equivalent to solving it once, except that the computation involved in the first task may be different from the one performed in the second instance.  Indeed, one of the results of running the SOAR architecture is the observation of the \emph{power law of practice} in successive duration measurements of consecutive executions of the same task \cite{Soar}.

In the NXP Architecture, full traces of the problem-solving activity are passed to the clustering subsystem in all cases, whether the final hypotheses are consistent or not with the initial expectation.  Although unexpected situations may occur---when final hypotheses and initial expectations are disjoint, for instance, and the degree to which they differ is an important input to clustering---both expected and unexpected result traces are used in clustering. In this respect, the semantics is best represented by continuations.

Following Hayo Thielecke's formal contrast between exception and continuations in stateful and stateless $\lambda$-calculus \cite{exncontstate}, exceptions are dynamic in nature while continuations are usually strongly associated with static binding.  Thielecke has shown that in the stateless case there are contextual equivalences that hold with exceptions and not with continuations (and vice versa).  Interestingly enough, the contextual equivalence $[[t]] \cong [[t \circ t]]$ outlined above actually holds even if both exceptions and continuations are present in the language in the absence of state.  A striking conclusion of this work is that one can draw a line between issues of control and issues of state.  The line, however, is fine as slight variants of continuation combined with exception can, in fact, express state.

In summary, the subtle differences in the relation between semantics of knowledge acquisition methods and problem-solving methods may lead to requirements of a combination of exception, continuation and state semantics.  The abstraction level we are interested in is less restrictive.  Not only these special requirements can be factored in more or less independently, as shown by Thielecke, but a continuation semantics can easily be defined for exceptions by passing two continuations as in the previous subsection.

\section{A Categorical View of Dynamic Memory}
Properties of continuations can be explored in more depth in a categorical context.  In a categorical framework the modularity of continuations is explicit and results from the categorical construct itself.

\subsection{Motivations}
Since the initial application of the concept of a \emph{triple}, or \emph{monad}, a categorical notion, to structure the denotational semantics of programming languages by Moggi \cite{moggi91notions}, several major results were derived regarding the semantics of both imperative and functional programming languages.  Control categories, for instance, are  used in the semantics of the lambda-mu calculus, an extension of the simply-typed lambda calculus with certain sequential control operators, to show the duality of \emph{call-by-name} and \emph{call-by-value} parameter passing techniques.  An interesting aspect of this duality is that it exchanges functional and imperative features \cite{selinger01control, filinski96controlling}.  This duality maps a purely functional call-by-value term to a call-by-name term that relies almost exclusively on control operators and vice versa.  This isomorphism of call-by-name and call-by-value, discovered by Filinski, finds in a categorical semantics a unified framework which abstracts away the actual syntactic issues.  In computational terms, this duality can be understood in a duality between data-driven and demand-driven computation, which reverses the direction of data---in proof-thoretical terms, it is an extension of the De Morgan duality from formula to proofs.

A common theme in the study of the operation of dynamic memory in cognitive systems is the \emph{complementarity} of performance and skill acquisition, as mentioned in the first section.  In this section, we borrow from category theory to model this complementarity in a framework of categorical semantics duality.  The introduction of bi-continuations to formalize learning as a side-effect of problem-solving in cognitive systems, as in the previous section, is a first step towards the introduction of an appropriate categorical framework for dynamic memory.  Following Moggi, Wadler \cite{wadler94monads, wadler93monads} elaborated on the deep relationship between continuations and monads.  In the  research work of Thielecke \cite{exncontstate}, Selinger \cite{selinger01control} and F\"{u}hrmann \cite{führmann00direct}, continuations are also linked to simpler categorical structures called control categories or premonoidal categories, which are central objects in the duality between call-by-name and call-by-value languages.  These categorical structures are also essential to the cognitive style of duality we are interested in here; we start with a brief review of some categorical background.

\subsection{Categorical background}
This section defines the basic notions from category theory that
we need in the formalisation of the skill acquisition side-effect in
monadic style. Readers are referred to \cite{ttt} for a
comprehensive presentation of categories and triples. Let $C$ be a
category, we denote by $Obj(C)$ the objects of $C$ and by
$Hom(A,B)$ the set of arrows with source object $A$ and target
object $B$.
\begin{defn} \label{d:functor}
If ${\bf C}$ and ${\bf D}$ are categories, a functor $F:{\bf C}
\rightarrow {\bf D}$ is a map for which:
\begin{itemize}
\item  If $f: A \rightarrow B$ is an arrow of ${\bf C}$, then $Ff:
FA \rightarrow FB$ is an arrow of ${\bf D}$; \item $F(id{_A}) =
id_{FA}$; and \item If $g: A \rightarrow B$, then $F(g \circ f) =
Fg \circ Ff$.
\end{itemize}
\end{defn}
A functor is a morphism of categories, a map which takes objects
to objects, arrows to arrows, and preserves source, target,
identities and composition. More generally $F$ {\em preserves} a
property P that an arrow $f$ may have if $F(f)$ has property P
whenever $f$ has. It {\em reflects} property P if $f$ has the
property whenever $F(f)$ has.

A natural transformation is defined as a "deformation" of one
functor to another.
\begin{defn}
If $F:{\bf C} \rightarrow {\bf D}$ and $G:{\bf C} \rightarrow {\bf
D}$ are two functors, $\lambda : F \rightarrow G$ is a natural
transformation from $F$ to $G$ if $\lambda$ is a collection of
arrows $\lambda C \rightarrow \lambda D$, one for each object $C$
of ${\bf C}$, such that for each arrow $g: C \rightarrow C'$ of
${\bf C}$ the following diagram commutes:

\def\ComponentOne{{\lambda C}}
\def\FuncOne{{Fg}}
\def\FuncTwo{{Gg}}
\def\ComponentTwo{{\lambda C'}}
\begin{diagram}
FC & \rTo^{\ComponentOne} & GC \\
\dTo<\FuncOne & & \dTo>\FuncTwo \\
FC' & \rTo^{\ComponentTwo} & GC'\\
\end{diagram}

The arrows $\lambda C$ are the components of $\lambda$. The
natural transformation $\lambda$ is a {\em natural equivalence} if
each component of $\lambda$ is an isomorphism in ${\bf D}$.
\end{defn}

\subsubsection{Triples, monads and categories for computations}
Triple or monads are, from one point of view, abstraction of
certain properties of algebraic structures, namely monoids. They
are categorical constructs that originally arose in homotopy
theory and were used in algebraic theory. Moggi
\cite{moggi91notions} was the first to discover the connection
between triples and semantics of effects in programming language
design. Since then the monadic style has pervaded theoretical
research on denotational and operational semantics.
\begin{defn} \label{d:triple}
A triple ${\bf T} = (T, \eta, \mu)$ on a category ${\bf C}$ is an
endofunctor $T: {\bf C} \rightarrow {\bf C}$ together with two
natural transformations $\eta: id{_{{\bf C}}} \rightarrow T$,
$\mu: TT \rightarrow T$ subject to the following commutative
diagrams:

\def\mut{{\mu T}}
\def\tmu{{T \mu}}
\begin{diagram}
TTT & \rTo^\mut & TT \\
\dTo<\tmu & {\rm associativity} & \dTo>{\mu} \\
TT & \rTo^{\mu} & T \\
\end{diagram}

expressing associative identity, and:

\def\etat{{\eta T}}
\def\teta{{T \eta}}
\begin{diagram}
T   & \rTo^\etat    & TT        & \lTo^\teta    & T \\
    & \rdTo<{=}     & \dTo>{\mu}& \ldTo>{=}     &   \\
    &               & T         &               &   \\
\end{diagram}

expressing left and right unitary identities.  The component of
$\mu T$ at an object $X$ is the component of $\mu$ at $TX$, while
the component of $T \mu$ at X is $T(\mu X)$; similar descriptions
apply to $\eta$.
\end{defn}

\begin{rem} \label{d:monad}
There is an alternate way of defining a triple based on a result
due to Manes.

Let ${\bf C}$ be a category with:
\begin{itemize}
\item A function $T : Obj(C) \rightarrow Obj(C)$; \item for each
pair of objects $C$ and $D$, a function $Hom(C,TD) \rightarrow
Hom(TC,TD)$, denoted $f \rightarrow f^{*}$; \item for each object
$C$ of ${\bf C}$ a morphism $\eta C : C \rightarrow TC$;
\end{itemize}
subject to the following conditions:
\begin{itemize}
\item For $f : C \rightarrow TD$, $f = \eta TD \circ f^{*}$; \item
for any object $C$, $(\eta C)^{*} = id_{TC}$; \item  for $f : C
\rightarrow TD$ and $g : D \rightarrow TE$, $(g^{*}\circ f)^{*} =
g^{*} \circ f^{*}$; \end{itemize} is equivalent to a triple on
${\bf C}$.
\end{rem}

The equivalence results from constructions of triples from adjoint
pairs separately discovered by Eilenberg-More and by Kleisli. The
function $ (\_)^{*}: Hom(C,TD) \rightarrow Hom(TC,TD)$ is also
known as the Kleisli star. This alternate definition emphasises
the connection between a triple and a monoid, an algebraic
structure with an associative operation and a unit element. Wadler
suggested a straightforward interpretation of the Kleisli star in
programming language semantics. In this context, the purpose of
the star operation is to combine two computations, where the
second computation may depend on a value yielded by the first
\cite{wadler94monads}. More precisely if $m$ is a computation of
type $T \tau_{1}$ and $k$ a function from values to computations
(such as a continuation) $\tau_{1} \rightarrow T\tau_{2}$, then
$k^{*}(m)$, or $m*k$, is of type $T\tau_{2}$ and represents the
computation that performs computation $m$, applies $k$ to the
value yielded by the computation, and then performs the
computations that results. It binds the result of computation $m$
in computation $k$. Different definitions for the triple $T$ and
the star operation then give rise to different monads to represent
different control operators such as {\tt escape/exit}, {\tt
call/cc}, {\tt prompt/control} or {\tt shift/reset}
\cite{wadler93monads,queinnec93library}.

As noted by Wadler and others, monadic and continuation-passing
styles appear closely related \cite{wadler94monads}. The actual
correspondence, however, is formally quite involved. Filinski has
shown the remarkable result that \emph{any} monadic effect whose
definition is itself expressible in a functional language can be
synthetised from just two constructs: first-class continuations
and a storage cell \cite{filinski94representing,
filinski96controlling}.  

Furthermore, Selinger and others \cite{selinger01control, führmann00direct} have used somewhat simpler categories than triples, the so-called \emph{pre-monoidal} categories, as minimal structures accounting for continuations.  These categorical structures were  successfully used to show the duality of call-by-value and call-by-name lambda calculus, a result previously obtained by Filinski.

\subsection{A categorical duality between performance and knowledge acquisition}

\subsubsection{Categories, computations and continuations}
Along the lines of the formalisation of the NXP architecture \cite{chauvet2002}, we represent the performance of a cognitive system as a goal-satisfaction computation.  In order to do so, we consider goals as expressions in a simple language--called the \emph{simple NXP language} in \cite{chauvet2002}--and the computation simply as the evaluation of these expressions by an appropriate interpreter.  Following Wadler's presentation style \cite{wadler93monads,wadler94monads} the interpreter evaluates an expression, of type $Exp$, and returns a result of type $T \; {\bf B}$ where ${\bf B}$ is a given output or response type and $T$, a triple or monad, is a type constructor capturing possible side-effects and exceptions.

The first operation defined in $T$ turns a value into a computation that returns that value and nothing else: $$\eta :: x \rightarrow T\;x$$
The second operation defined in $T$ simply applies a function of type $x \rightarrow T \; y$ to a computation of type $T\;x$.  Following again Wadler's notation we write the argument before the function: $$(*) :: T\;x \rightarrow (x \rightarrow T\;y) \rightarrow T\;y$$

From a computation perspective, the categorical laws defining the triples $(T,\eta,*)$ translate to:
\begin{defn}\label{d:TripleOperations}
\emph{Left unit operation}: Compute the value $x$, bind $y$ to the result and compute $z$; the result is the same as $z$ with value $x$ substituted for variable $y$.$$ \eta x \;*\; \lambda y.z = z\left[x/y\right]$$

\emph{Right unit operation}: Compute $x$, bind the result to $y$ and returns $y$; the result is the same as $x$.$$x\;*\;\lambda y.\eta y = x$$

\emph{Associativity}: Compute $x$, bind the result to $y$, compute $z$, bind the result to $t$ and compute $o$; the order of parentheses is irrelevant.$$x\;*\;(\lambda y.z\;*\;\lambda t.o) = (x\;*\;\lambda y.z)\;*\;\lambda t.o$$
\end{defn}
\begin{defn}\label{d:Evaluator}
The performance of a goal-satisfaction cognitive system is defined by the execution of an interpreter $eval$ defined by as follows for the simple NXP language :
$$ b \;|\; E\;and\;E \;|\; E\;or\;E $$
with $b$ in ${\bf B}$ and $E$ in $Exp$.

and

\begin{align*}
eval &: Exp \rightarrow T {\bf B}  \\
eval(b) &= \eta b \\
eval(E_{1}\;or\;E_{2}) &= eval(E_{1})*\lambda
x.eval(E_{2})*\lambda y.\eta(x\;|\;y) \\
eval(E_{1}\;and\;E_{2}) &= eval(E_{1})*\lambda
x.eval(E_{2})*\lambda y.\eta(x\;\&\;y) \\
\end{align*}
\end{defn}

\subsubsection{Postponed evaluation}
In \cite{chauvet2002} goal evocation is outlined as a critical process in the performance of a goal-based cognitive system.  In the categorical framework, the semantics of goal evocation is simply represented by an additional operation, $post$, defined on the computational triple above.

In the new triple, a computation accepts an initial memory state and returns a value, as before, paired with a final state.  Definitions of the unit and the star operations are revised to accommodate the new types thus defined.

\begin{defn}\label{d:TripleWithState}
The memory triple is defined by $(T,\eta,*)$ with:
\begin{align*}
T\;x &= Mem \rightarrow (x,Mem) \\
\eta &:: x \rightarrow T\;x \\
\eta x &= \lambda s.(x,s) \\
(*) &:: T\;x \rightarrow (x \rightarrow T\;y) \rightarrow T\;y \\
x\;*\;k = \lambda s.&{\bf let} (y,a)\;=\;x\;s\;{\bf in} \\
 &{\bf let} (z,b)\;=\;k\;x\;a\;{\bf in} \\
 &(z,b) \\
\end{align*}
\end{defn}

The goal evocation is introduced by specifying the $Mem$ set of cognitive states as the set of \emph{sequences} of goals and subgoals \cite{chauvet2002}, and defining $post$ only as a side effect on states.

\begin{defn}\label{d:TripleWithEvocation}
The NXP triple, or $N$-triple, is a triple with state as in \ref{d:TripleWithState} where $Mem$ is the set $Exp^{*}$ of sequences of elements of $Exp$ (with its appropriate internal operations) and a function $post$:
$$post\; E = \lambda s.((),s + {\tt inr}(E))$$
where {\tt inr} is the right injection ${\bf B}\times {\bf B}^{*}
\rightarrow {\bf B}^{*}$.
\end{defn}

The evaluator is also revised to take the new function into account by adding the rule:$$eval(b \;post\; E) = post E * \eta b$$
as expected.  In computational terms, the evaluation of an expression may result in posting one or several additional expressions for later evaluations.  The categorical framework captures the delayed execution as an on-the-fly modification to the current continuation.  $N$-triples can be viewed as traditional continuations categories with this additional continuation function built-in.

\subsection{Categorical skill acquisition}
In this section we bring skill acquisition and performance of cognitive systems under the same mathematical representation, namely a specific continuation category, the $N$-triple introduced in \ref{d:TripleWithEvocation}.  A practical result of this identity is a suggestion for a very broad implementation architecture for both problem-solving and skill acquisition threads in a computing process.

The semantics expressed by the categorical definition \ref{d:TripleWithEvocation} of $N$-triples captures the central idea that a side-effect of a computing process could simply be to trigger delayed execution of additional computing, once the current process is exhausted.  This \emph{abstract} side-effect may be put to use in different settings.  Focusing on the performance aspect of the problem-soving activity, the $N$-triple induced semantics represents the \emph{goal-evocation} mechanism as in \cite{chauvet2002}.  In this instance, as problem-solving progresses, further goals are posted for later evaluation once the current goal investigation terminates.  Focusing on the skill acquisiton aspect of cognitive systems, the semantics of $N$-triples expresses that performance induces a state of expectation which improves later problem-solving tasks.  As the cognitive system solves problems, experience builds up which improves further problem solving.  From a computing perspective, the model is of skill acquisition as a side-effect of problem solving.

Furthermore, it is important to note that there is a difference in the way the $N$-triple side-effect is used in the first instance and in the second one.  At the problem-solving level, the intent is that the evoked goals are determined from an examination of the syntax of the goals and subgoal trees.  In the NXP architecture of \cite{chauvet2002}, evaluation of a subgoal triggers the posting of all other subgoals sharing the boolean variable in the current subgoal syntactic expression (in the simplified NXP language).  These dependencies can in fact be compiled beforehand into an association network structure used at runtime to simulate the goal-evocation process.  In the the skill acquisition study, however, the state is made a of collection of expressions to be evaluated once the current problem-solving case is solved.  This state is made dependent on the sequential history of all previous problem-solving cases and cannot be determined from the simple examination of the syntactic representation of the knowledge perused in these problem-solving activities.  This state of \emph{expectation} results from the experience of the cognitive agent.

In contrast to the the static nature of the dependency network in the problem-solving process representation, skill acquisition then relies on a dynamic dependency network built on the fly as the sequence of cases are being solved.  In particular, several models of construction of the expectation state can readily be \emph{plugged in} into the concrete semantics when working towards an implementation of an $N$-triple virtual machine.

The stack-based implementation semantics proposed in \cite{chauvet2002} may be reused here to factor in skill acquisition, in various flavors, into the NXP architecture.  Informally, this implementation semantics uses stacks of subgoals to represent problem-solving as a computing process.  As a goal requires evaluation of one or several subgoals, the latter are pushed on the stack.  As a goal is evoked during evaluation of the current one it is added at the bottom of the stack.  The concrete implementation semantics simply describes the proper alternance of stack operations (pusp, pop and add) as problem solving proceeds.

An enriched stack based implementation semantics is suggested here as a substitution to this one in order to complete it with skill acquisition as a side-effect.  Informally again, the new implementation semantics makes use of two different stacks.  The first one, the problem-solving stack, is used as before.  The second one, the skill acquisition stack, stores expressions constituting the expectation state as the problem solving proceeds.  These expressions are the result of a given \emph{learning} function being applied to the current problem-solving stack and the skill acquisition stack itself : ${\tt learn}\;:S \times S \rightarrow S$.

This general extension encompasses both the cases where learning happens during the solving activity as in Schank and Abelson's dynamic memory models and in the SOAR architecture, and those where acquisition happens in discrete steps at the end of each problem-solving episode as in the NXP models \cite{Rappaport_1984_77}.  From an implementation perspective this concrete semantics is easily mapped to another lower-level concrete semantics that peruses only one stack to represent both the problem-solving one and the skill acquisition one.  This, in fact, is the original stack-based concrete semantics of \cite{chauvet2002} with additional ancillary stack operations (such as merge, for instance, concatenating two stacks either at the top or at the bottom).  This broad implementation semantics defines a $N$-triple virtual machine displaying coprocesses of interdependent problem-solving and skill acquisition.

\section{Conclusion and further work}
The categorical framework simplifies the formal study of cognitive processes, such as problem-solving, search and learning, as specific types of computing processes.  Within category theory, programming languages constructs and execution architecture find a natural, mathematical expression.  Side-effects, in particular, are now well understood in the categorical framework.  Results in category theory nicely translate to properties of computing processes and may even guide the design of programming languages tools such as compilers, for instance.  The Glasgow implementation of the Haskell programming language makes extensive use of triples, a categorical notion, a design which was further adopted for version 1.3 of the Haskell standard.

Categories also help formalizing system-level behaviors of cognitive agents in the A.I. framework.  The intuition behind this categorical formal step is to capture these behavioral properties as side-effects of the cognitive process considered as a computing process.  In a companion paper \cite{chauvet2002} a simple programming language was introduced to represent a (simplified) cognitive process, problem-solving or search in a typical Simon-Newell framework \cite{Newell93}.  The main result of that paper was to propose a categorical view of elementary operations of a cognitive agent or system, subgoal evaluation and evocation.  This categorical approach informed the design of the stepwise refinement of an abstract semantics into an implementation concrete semantics, called the NXP architecture.

This paper revisited the NXP architecture and investigated the relevance of the categorical framework in the study of learning and, more specifically, of knowledge and skill acquisition.  Abstracting away from several well-established models, it was shown that category theory provided indeed formal tools to help in the understanding of skill acquisition and its correlation to problem-solving and search.  It introduced the $N$-triple as a simple extension of the triple or monad, to capture the notion of expectation state.

Finally, the implementation semantics introduced in the first paper was shown to be applicable to the enriched semantics induced by the $N$-triple construction.  By introducing an intermediary double-stack based concrete semantics, which ultimately resolves into the original single-stack semantics, it was then suggested that the NXP architecture exhibits the dual nature of problem-solving and skill or knowledge acquisition.  From an implementation perspective, the NXP architecture defines a virtual machine for a cognitive agent or system, in which problem-solving and knowledge or skill acquisition co-occur as it confront its environment.  Considering cognitive behaviors as computing processes---and formalizing this intuition---opens up new avenues for optimal implementation of artificial agents exhibiting intelligent behavior.

\bibliographystyle{plain}
\bibliography{wip}

\begin{thebibliography}{10}

\bibitem{Barendregt-1997}
Hendrik~P. Barendregt.
\newblock The impact of the lambda calculus on logic and computer science.
\newblock {\em Bulletin of Symbolic Logic}, 3(3):181--215, 1997.

\bibitem{ttt}
Michael Barr and Charles Wells.
\newblock {\em Toposes, Triples and Theories}, volume 278 of {\em Grundlehren
  der mathematischen Wissenschaften}.
\newblock New York, 1985.
\newblock A list of corrections and additions is maintained in \cite{tttcorr}.

\bibitem{tttcorr}
Michael Barr and Charles Wells.
\newblock Corrections to {T}oposes, {T}riples and {T}heories.
\newblock Available by anonymous FTP from {\tt triples.math.mcgill.ca} in
  directory {\tt pub/barr}, 1993.
\newblock Corrections and additions to \cite{ttt}.

\bibitem{chauvet2002}
Jean-Marie Chauvet.
\newblock Monadic style control constructs for inference systems.
\newblock Technical Report arXiv.org ePrint Archive cs.AI/0211035, Dassault
  D\'{e}veloppement, Paris, Fr, August 2002.

\bibitem{führmann00direct}
C.~Führmann.
\newblock Direct models for the computational lambda calculus.
\newblock In Stephen Brookes, Achim Jung, Michael Mislove, and Andre Scedrov,
  editors, {\em Electronic Notes in Theoretical Computer Science}, volume~20.
  Elsevier, 2000.

\bibitem{filinski94representing}
Andrzej Filinski.
\newblock Representing monads.
\newblock In {\em Conf.\ Record 21st {ACM} {SIGPLAN}-{SIGACT} Symp.\ on
  Principles of Programming Languages, {POPL}'94, Portland, {OR}, {USA}, 17--21
  Jan.\ 1994}, pages 446--457. ACM Press, New York, 1994.

\bibitem{filinski96controlling}
Andrzej Filinski.
\newblock {\em {C}ontrolling {E}ffects}.
\newblock PhD thesis, Pittsburgh, Pennsylvania, May 1996.

\bibitem{27702}
John~E. Laird, Allen Newell, and Paul~S. Rosenbloom.
\newblock Soar: an architecture for general intelligence.
\newblock {\em Artificial Intelligence}, 33(1):1--64, 1987.

\bibitem{moggi91notions}
Eugenio Moggi.
\newblock Notions of computation and monads.
\newblock {\em Information and Computation}, 93(1):55--92, 1991.

\bibitem{Newell93}
A.~Newell.
\newblock The knowledge level.
\newblock In P.~S. Rosenbloom, J.~E. Laird, and A.~Newell, editors, {\em The
  Soar Papers: Research on Integrated Intelligence (Volume 1)}, pages 136--176.
  MIT Press, London, 1993.

\bibitem{Soar}
Allen Newell.
\newblock {\em Unified Theories of Cognition}.
\newblock Harvard University Press, 1990.

\bibitem{queinnec93library}
Christian Queinnec.
\newblock A library of high-level control operators.
\newblock 6(4):11--26, 1993.

\bibitem{Rappaport_1984_77}
Alain Rappaport and Jean-Marie~C. Chauvet.
\newblock Symbolic knowledge processing for he acquisition of expert behavior:
  A study in medicine.
\newblock Technical Report CMU-RI-TR-84-08, Robotics Institute, Carnegie Mellon
  University, Pittsburgh, PA, May 1984.

\bibitem{scripts}
Robert~Abelson Roger~Schank.
\newblock {\em Script Plans Goals and Understanding}.
\newblock Lawrence Erlbaum Associates, 1977.

\bibitem{sabry92reasoning}
Amr Sabry and Matthias Felleisen.
\newblock Reasoning about programs in continuation-passing style.
\newblock In {\em Proceedings 1992 {ACM} Conf.\ on Lisp and Functional
  Programming, San Francisco, {CA}, {USA}, 22--24 June 1992}, pages 288--298.
  ACM Press, New York, 1992.

\bibitem{DM}
Roger Schank.
\newblock {\em Dynamic Memory}.
\newblock Cambridge University Press, 1982.

\bibitem{selinger01control}
Peter Selinger.
\newblock Control categories and duality: On the categorical semantics of the
  lambda-mu calculus.
\newblock {\em Mathematical Structures in Computer Science}, 11(2):207--260,
  2001.

\bibitem{Stoy}
J.~E. Stoy.
\newblock {\em Denotational Semantics: The Scott-Strachey Approach to
  Programming Languages}.
\newblock MIT Press, Cambridge, Mass., 1977.

\bibitem{exncontstate}
Hayo Thielecke.
\newblock On exceptions versus continuations in the presence of state.
\newblock In Gert Smolka, editor, {\em Programming Languages and Systems, 9th
  European Symposium on Programming, ESOP 2000}, number 1782 in LNCS, pages
  397--411. Springer Verlag, 2000.

\bibitem{wadler93monads}
Philip Wadler.
\newblock Monads for functional programming.
\newblock In M.~Broy, editor, {\em Program Design Calculi: Proceedings of the
  1992 {M}arktoberdorf International Summer School}. Springer-Verlag, 1993.

\bibitem{wadler94monads}
Philip Wadler.
\newblock Monads and composable continuations.
\newblock {\em Lisp and Symbolic Computation}, 7(1):39--56, 1994.

\bibitem{PDIS}
D.~Waterman and Eds. F.~Hayes-Roth.
\newblock {\em Pattern- Directed Inference Systems}.
\newblock Academic Press, New York, NY, 1978.

\end{thebibliography}
\end{document}